\documentclass{article}

\usepackage{booktabs}
\usepackage[final]{corl_2026} 
\usepackage{graphicx}
\usepackage{amsmath}
\usepackage{amssymb}
\usepackage{hyperref}

\hypersetup{
    pdftitle={Humanoid Badminton: Learning Dynamic Racket Skills from Limited Human Motion Data},
    pdfauthor={Jingzhi Cui and Zhexiong Wang and Bangjie Xu and Pengyu Zhao and Youyuan Li and Zhi Su and Peng Ren and Mengdi Xu and Chao Yu and Yi Wu and Luyang Wang and Zhongyu Li},
    pdfkeywords={Humanoid, Athletic Robotics, Reinforcement Learning}
}

\title{Humanoid Badminton: Learning Dynamic Racket Skills from Limited Human Motion Data}

\author{
\textbf{
  Jingzhi Cui\textsuperscript{1, 3}, 
  Zhexiong Wang\textsuperscript{3, 5}, 
  Bangjie Xu\textsuperscript{1}, 
  Pengyu Zhao\textsuperscript{4}, 
  Youyuan Li\textsuperscript{1}, 
  }\\\textbf{
  Zhi Su\textsuperscript{1}, 
  Peng Ren\textsuperscript{6}, 
  Mengdi Xu\textsuperscript{1}, 
  Chao Yu\textsuperscript{1}, 
  Yi Wu\textsuperscript{1}, 
  Luyang Wang\textsuperscript{4}, 
  Zhongyu Li\textsuperscript{2, 3}
  }
  \\[2mm]
  \textsuperscript{1}Tsinghua University, 
  \textsuperscript{2}The Chinese University of Hong Kong, 
  \textsuperscript{3}Hong Kong Embodied AI Lab, \\
  \textsuperscript{4}Beijing University of Civil Engineering and Architecture, 
  \textsuperscript{5}Zhejiang University, 
  \textsuperscript{6}DeepCybo
  \\[2mm]
  Emails: \texttt{cjz24@mails.tsinghua.edu.cn, zhongyuli@cuhk.edu.hk} 
  \\
  Website: \url{https://sunlight02.github.io/humanoid-badminton/}
}

\begin{document}
\maketitle


\begin{figure}[h]
    \centering
    \includegraphics[
        width=1.0\linewidth,
        trim=170 225 160 225,
        clip
    ]{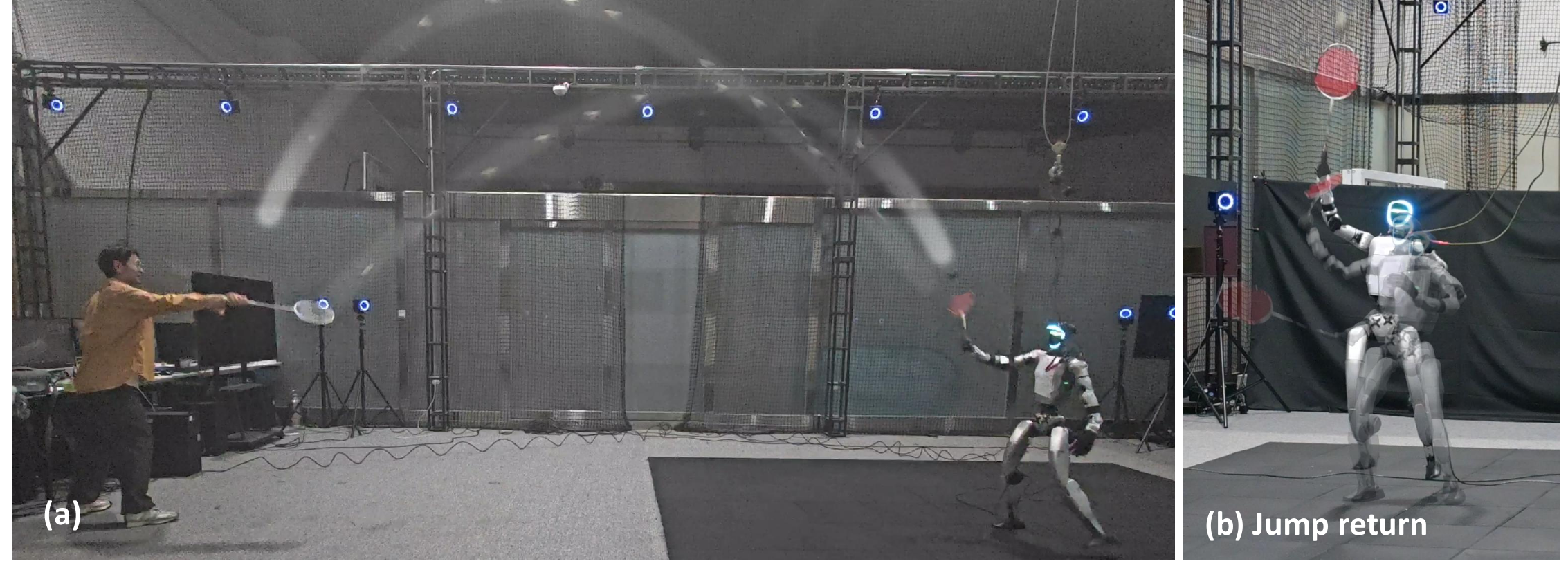}
    \caption{
    \textbf{Real-world humanoid badminton.}
    (a) Human--robot rallies require timely decisions, coordinated footwork and racket control, and whole-body balance.
    (b) A highly dynamic jump return combines rapid takeoff, aerial striking with a high-speed racket swing, and stable landing.
    }
    \label{fig:intro_overview}
\end{figure}

\begin{abstract}
High-speed racket sports provide a demanding testbed for humanoid robots, requiring time-critical decisions, precise striking, and dynamic whole-body coordination.
In badminton, fast-changing shuttle trajectories require timely contact decisions, while successful returns demand precise racket pose and velocity within a brief contact window and across a broad three-dimensional striking workspace.
Human motion data provide valuable priors for such athletic skills, but usable badminton references are limited and imperfect.
Direct tracking provides insufficient executable variation for diverse shuttle conditions, while purely task-driven optimization may produce unnatural motion.
We therefore present a three-stage hierarchical reinforcement learning framework for dynamic humanoid badminton.
First, task-randomized motion augmentation expands sparse annotated hitting events into executable target-conditioned stroke variations, forming a continuous latent skill space.
Second, a high-level planner outputs continuous latent skill codes to compose these skills online according to the observed shuttle state.
Third, a context-conditioned adversarial regularizer encourages more natural planner-level skill usage while preserving return performance.
In real-world deployment, our system sustains multi-skill rallies with human players, including forehand, backhand, and highly dynamic jump returns.
This is the first real-world humanoid racket-sport system to demonstrate multi-skill human--robot rallies including highly dynamic jump returns.
\end{abstract}



\section{Introduction}

Recent advances in humanoid locomotion and whole-body control have enabled robots to perform increasingly dynamic behaviors~\citep{Cheng2024ExBody,Zhang2024WholeBodyHumanReference,He2025ASAP,liao2026beyondmimic}.
Whereas many commonly studied humanoid tasks center on stable locomotion or relatively slow object interactions, athletic tasks require robots to react to fast-moving objects, make time-critical decisions, generate precise contacts, and maintain balance during dynamic whole-body motion.
High-speed racket sports naturally expose these challenges, with recent progress in table tennis~\citep{Dambrosio2024TableTennis,Su2025Hitter}, tennis~\citep{Terasawa2016HumanoidTennis,Hattori2020HumanoidTennis,Zaidi2023WheelchairTennis,Zhang2023Vid2Player3D,Zhang2026Latent}, and badminton on both quadrupedal platforms~\citep{Ma2025LeggedBadminton} and humanoid robots~\citep{Liu2025HumanoidBadminton,chen2026learninghumanlikebadmintonskills}.

Badminton presents a distinctive setting for humanoid robots.
The lightweight shuttlecock is strongly affected by aerodynamic drag, resulting in fast-changing trajectories.
At the same time, successful returns require precise racket--shuttle contact: the robot must select a feasible contact point within a broad three-dimensional striking workspace and coordinate the racket position, orientation, and velocity to intercept and redirect the shuttlecock within a brief contact window.
Compared with quadrupedal platforms, humanoid robots must coordinate large-amplitude racket swings with rapid whole-body motion while maintaining balance with a limited support region and a relatively high center of mass.
These challenges become particularly pronounced in highly dynamic jump returns, which involve rapid takeoff, aerial striking, and stable landing. 
Recent humanoid badminton systems have demonstrated real-world shuttle interception and dynamic whole-body striking~\citep{Liu2025HumanoidBadminton,chen2026learninghumanlikebadmintonskills}.
However, sustained multi-skill rallies over a broad striking workspace, particularly those involving highly dynamic skills such as jump returns, remain underexplored.

Purely task-driven reinforcement learning (RL) provides a direct way to optimize badminton performance, but unconstrained task optimization may produce jittery or unnatural motions.
Human motion data provide useful priors for natural athletic motion, yet usable badminton references are scarce and remain imperfect after video reconstruction and cross-embodiment retargeting.
Moreover, tracking a small set of references provides insufficient variation to accommodate diverse shuttle conditions.
This motivates expanding limited human references into executable stroke variations while preserving their motion structure.
A hierarchical skill interface can combine these motion priors with task-driven optimization.
However, a high-level planner optimized primarily for return success may use even well-learned low-level skills in atypical ways, leading to less natural motion.
This motivates regularizing planner-level latent skill usage while preserving task performance.

To address these challenges, we propose a three-stage hierarchical RL framework for dynamic humanoid badminton.
First, we construct a continuous target-conditioned latent skill space from reconstructed and retargeted badminton motions through task-randomized motion augmentation.
For each annotated hitting event, diverse racket targets define randomized hitting tasks, and RL-based adaptation expands the corresponding reference motion into executable local stroke variations.
Second, we train a high-level planner to map shuttle observations and robot state to continuous latent skill codes, enabling online composition of the learned skills for dynamic returns.
Third, we introduce a context-conditioned adversarial regularizer that encourages more natural planner-level skill usage while preserving return performance.
When deployed on a real humanoid robot, the learned policy achieves a high return success rate while sustaining multi-skill rallies with human players, including forehand, backhand, and highly dynamic jump returns.
This is the first real-world humanoid racket-sport system to demonstrate multi-skill human--robot rallies including highly dynamic jump returns.

Our main contributions are as follows:
\noindent\textbf{(1) A staged hierarchical learning framework for dynamic humanoid badminton} that learns diverse racket skills from limited and imperfect human motion data.
\noindent\textbf{(2) A task-randomized motion augmentation strategy} that expands sparse annotated hitting events into executable target-conditioned stroke variations through randomized hitting tasks and RL-based adaptation.
\noindent\textbf{(3) A context-conditioned adversarial regularizer for hierarchical RL} that encourages more natural planner-level skill usage while preserving task performance.
\noindent\textbf{(4) The first real-world humanoid racket-sport system to demonstrate multi-skill human--robot rallies with highly dynamic jump returns.}


\section{Related Work}
\label{sec:related_work}
We review related work on robotic racket sports, reference motion generation and augmentation, and hierarchical skill learning and regularization.

\noindent\textbf{Robotic racket sports.}
Prior work has explored robotic table tennis for competitive play against human opponents~\citep{Dambrosio2024TableTennis} and humanoid whole-body control~\citep{Su2025Hitter}, as well as robotic tennis for dynamic humanoid swings~\citep{Terasawa2016HumanoidTennis,Hattori2020HumanoidTennis} and athletic skill learning from imperfect human motion data~\citep{Zhang2026Latent}.
For badminton, Ma et al.~\citep{Ma2025LeggedBadminton} demonstrated coordinated shuttle prediction, locomotion, and racket swinging on a quadrupedal mobile-manipulation platform.
Liu et al.~\citep{Liu2025HumanoidBadminton} developed a staged reinforcement learning framework enabling dynamic humanoid whole-body striking and real-world human--robot rallies, although their demonstrated interception region and stroke diversity remain relatively limited.
Chen et al.~\citep{chen2026learninghumanlikebadmintonskills} introduced an imitation-to-interaction framework with human motion priors, demonstrating forehand and backhand lift interception on hardware in a localized, single-shot setting.
Despite these advances, sustained humanoid badminton rallies combining broad-workspace interception, diverse racket skills, and highly dynamic jump returns remain underexplored.
Our framework addresses this setting by learning dynamic racket skills from limited human motion data and enabling sustained multi-skill human--robot rallies, including forehand, backhand, and jump returns.

\noindent\textbf{Reference motion generation and augmentation.}
Human motion data provide useful priors for coordinated whole-body skills through motion imitation and human-to-humanoid retargeting~\citep{Peng2018DeepMimic,Zhang2024WholeBodyHumanReference,Araujo2025GMR}.
However, reconstructed and retargeted motions may contain artifacts, and a limited set of references cannot cover the continuous variations in racket contact required for badminton.
Data augmentation offers a way to expand limited demonstrations, as exemplified by MimicGen in robotic manipulation~\citep{Mandlekar2023MimicGen}.
Task randomization has also improved robustness and generalization in bipedal locomotion~\citep{li2025reinforcement}.
Motivated by these approaches, we introduce task-randomized motion augmentation to learn executable stroke variations from limited human references.

\noindent\textbf{Hierarchical skill learning and usage regularization.}
Hierarchical control separates low-level motion execution from high-level task decisions and has been applied to dynamic tasks such as quadrupedal goalkeeping~\citep{huang2023creating} and humanoid table tennis~\citep{Su2025Hitter}.
Reusable latent skill representations can also be learned by distilling motion imitators~\citep{luo2024universal} or through adversarial learning from motion data~\citep{Peng2022ASE}.
However, a useful low-level motion prior does not guarantee that a high-level planner will use the learned skills appropriately across robot contexts.
Zhang et al.~\citep{Zhang2026Latent} introduced a latent action barrier that explicitly bounds high-level latent exploration around a state-dependent skill prior through a $\tanh$-based transformation.
In contrast, our framework uses a learned context-conditioned adversarial reward to encourage appropriate planner-level skill usage without imposing explicit latent-action bounds, while preserving task performance.


\section{Method}

We introduce a three-stage hierarchical framework that constructs a continuous target-conditioned latent skill space through task-randomized motion augmentation, trains a high-level planner for shuttle returns, and regularizes planner-level latent skill usage through a context-conditioned adversarial objective.

\begin{figure}[t]
\centering
\includegraphics[width=0.9\textwidth]{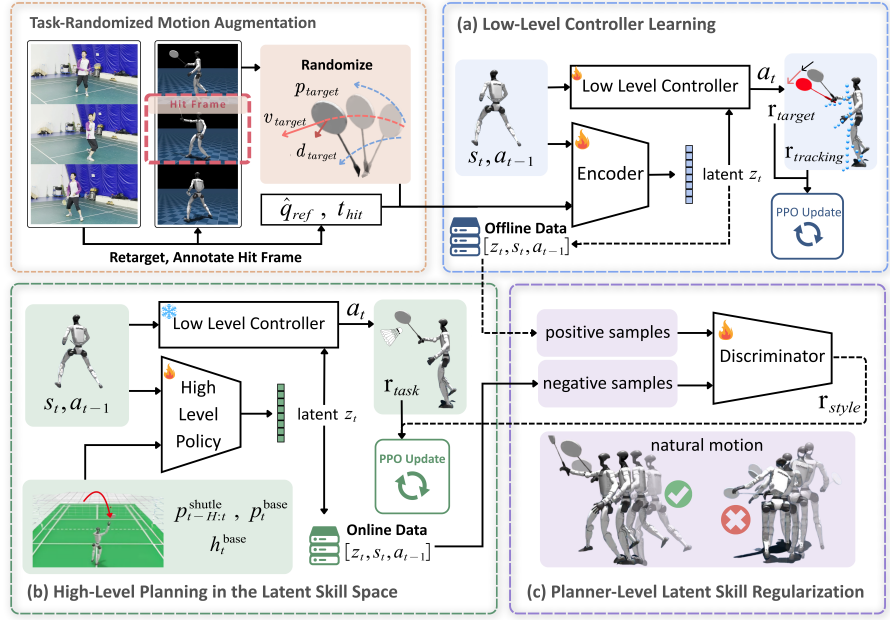}
\caption{
\textbf{Overview of the proposed hierarchical learning framework.}
We first reconstruct and retarget human badminton videos, annotate hitting events, and perform \textbf{task-randomized motion augmentation} by sampling racket targets for each annotated hitting event, including contact position $p_{\mathrm{target}}$, racket velocity $v_{\mathrm{target}}$, and racket-face direction $d_{\mathrm{target}}$.
In \textbf{(a)}, a target-conditioned skill encoder maps reference-motion information, the sampled target, hit timing, and robot context to a latent skill code $z_t$, while a low-level controller $\pi_\phi$ maps the latent code and robot context to joint actions.
In \textbf{(b)}, the low-level controller is frozen, and a high-level planner $\pi_\theta$ observes shuttle history and robot state information to output continuous latent codes for shuttle returns.
In \textbf{(c)}, a context-conditioned discriminator distinguishes Stage~1-generated latent--context pairs from planner-generated pairs, providing a latent-style reward for regularizing planner-level skill usage.
}
\label{fig:method_overview}
\end{figure}

As illustrated in Fig.~\ref{fig:method_overview}, our framework separates skill acquisition, task optimization, and planner-level latent skill regularization.
In Stage~1, we jointly train a target-conditioned skill encoder and a low-level controller from retargeted human motions.
For each annotated hitting event, diverse racket targets are sampled online to define randomized hitting tasks.
Through RL-based adaptation to these tasks, Stage~1 expands sparse hitting events into executable stroke variations and constructs a reusable continuous latent skill space.
In Stage~2, the low-level controller is frozen, and a high-level planner is trained to output latent skill codes for shuttle returns.
The planner therefore optimizes the badminton return task in the learned skill space rather than directly in the joint-action space.
In Stage~3, we continue planner training with a context-conditioned latent-style reward that encourages the planner to use latent skills in contexts consistent with the executable skill distribution learned in Stage~1, while preserving the task competence acquired in Stage~2.

\subsection{Stage 1: Low-Level Controller Learning with Task-Randomized Motion Augmentation}

Stage~1 learns a target-conditioned continuous skill space from limited retargeted human badminton motions.
We jointly train a skill encoder and a low-level controller through reference-motion tracking and randomized hitting tasks constructed from annotated hitting events.

\paragraph{Motion Processing.}
We use approximately 30 minutes of human badminton videos, divide them into short motion segments, reconstruct the human motions using GVHMR~\citep{Shen2024GVHMR}, and retarget them to the target humanoid embodiment using GMR~\citep{Araujo2025GMR}, yielding reference joint trajectories $q_{\mathrm{ref}}$.
For each hitting event $k$, we annotate its hit frame $t_k^*$ and define a local hit window $\mathcal{W}_k=[t_k^*-\delta,t_k^*+\delta]$, with $\mathcal{W}=\bigcup_k \mathcal{W}_k$ denoting the union of all hit windows in a segment.
We set $\delta=2$, corresponding to a five-frame window.

\paragraph{Task-Randomized Motion Augmentation.}
Directly tracking the retargeted motions exposes the controller to only a limited set of hitting conditions.
For each annotated hitting event, we sample a racket target $\tau_{\mathrm{target}}=(p_{\mathrm{target}},v_{\mathrm{target}},d_{\mathrm{target}})$, where $p_{\mathrm{target}},v_{\mathrm{target}}\in\mathbb{R}^3$ denote the desired contact position and racket velocity, and $d_{\mathrm{target}}\in\mathbb{S}^2$ denotes the desired racket-face direction.
During reinforcement learning, target racket states are sampled online for each event using the procedure detailed in Appendix~\ref{app:motion_processing}, while the corresponding reference motion and hit timing are retained.
Through RL-based adaptation to these randomized objectives, each annotated hitting event is expanded into a local family of executable target-conditioned stroke variations.

\paragraph{Latent Skill Encoding and Controller Learning.}
Let $s_t$ denote the robot proprioceptive state and $a_{t-1}$ the previous action, and define the robot context as $\tilde{s}_t=[s_t,a_{t-1}]$.
At each timestep, the skill encoder receives the reference-motion information, sampled target, time-to-hit, and robot context, and produces an $8$-dimensional latent skill code $z_t\in\mathcal{Z}=[-1,1]^8$ through a $\tanh$ output layer.
The low-level controller then maps the latent code and robot context to joint actions:
$$
a_t=\pi_\phi(z_t,\tilde{s}_t).
$$
The skill encoder and low-level controller are jointly optimized with PPO \citep{Schulman2017PPO} to realize the randomized hitting tasks.
The reward encourages reference-motion tracking outside the hit window and target matching inside the hit window:
\begin{equation}
r_t =
\mathbb{I}[t \notin \mathcal{W}] r_t^{\mathrm{track}}
+
\mathbb{I}[t \in \mathcal{W}] r_t^{\mathrm{target}}
+
w_{\mathrm{reg}} r_t^{\mathrm{reg}},
\end{equation}
where $\mathbb{I}[\cdot]$ is the indicator function.
Here, $r_t^{\mathrm{track}}$ rewards whole-body reference-motion tracking,
$r_t^{\mathrm{target}}$ rewards racket position, velocity, and face-direction matching relative to the sampled target,
and $r_t^{\mathrm{reg}}$ denotes regularization terms that promote stable and physically feasible motion.
Together, the skill encoder and low-level controller provide a compact latent interface for generating executable target-conditioned stroke variations, enabling downstream planning.

\subsection{Stage 2: High-Level Planning in the Latent Skill Space}

Stage~2 trains a high-level planner over the learned latent skill space $\mathcal{Z}$ for shuttle returns, while keeping the Stage~1 low-level controller frozen.
At each timestep, the planner observes a task-level state $o_t$ and outputs a latent skill code
\[
z_t \sim \pi_\theta(\cdot \mid o_t), \qquad z_t\in\mathcal{Z}.
\]
The resulting code is executed by the frozen low-level controller as $a_t=\pi_\phi(z_t,\tilde{s}_t)$.
This hierarchical interface allows the planner to optimize the badminton return task in the learned skill space rather than directly in the joint-action space.
The planner observation $o_t$ comprises the robot context $\tilde{s}_t$, the recent shuttle positions $p^{\mathrm{shuttle}}_{t-H+1:t}$, the robot base position $p_t^{\mathrm{base}}\in\mathbb{R}^3$, and the base heading direction $h_t^{\mathrm{base}}$, represented as a two-dimensional unit vector.
We use $H=5$ recent shuttle positions to characterize the incoming trajectory, while the base position and heading provide the robot's court-relative configuration.
The planner is trained with PPO under randomized shuttle launches from the opponent side; additional training details and reward groups are provided in Appendix~\ref{app:planner_training}.

\subsection{Stage 3: Planner-Level Latent Skill Regularization}

Stage~2 focuses on task performance but does not explicitly constrain how the planner uses latent skills across robot contexts.
To regularize planner-level skill usage, we introduce a context-conditioned discriminator $D_\psi$ over latent--context pairs $(z_t,\tilde{s}_t)$.
Positive samples are collected from Stage~1 rollouts under randomized hitting tasks, where latent codes are generated by the skill encoder and executed by the low-level controller, while negative samples are generated online by the planner.
The discriminator learns to distinguish executable latent--context usage induced by Stage~1 from planner-generated usage.
Its output defines a planner-level latent-style reward
$$
r_t^{\mathrm{style}}
=
-\log\left(1-D_\psi(z_t,\tilde{s}_t)\right).
$$
Starting from the Stage~2 planner checkpoint, we continue PPO training with
$$
r_t^{\mathrm{total}}
=
r_t^{\mathrm{task}}
+
w_{\mathrm{style}}r_t^{\mathrm{style}},
$$
where $r_t^{\mathrm{task}}$ is the Stage~2 task reward and $w_{\mathrm{style}}$ balances task performance and latent-style regularization.
The low-level controller $\pi_\phi$ remains frozen throughout this stage.
This regularization promotes more natural planner-level skill usage by encouraging latent skills to be used in contexts consistent with the executable skill distribution learned in Stage~1, while preserving the task competence acquired in Stage~2.

\section{Experiments}

We evaluate the proposed framework in simulation and real-world deployment. Our experiments address four questions:
\textbf{(1)} whether the proposed method improves task performance, stroke diversity, and motion quality over representative baselines;
\textbf{(2)} whether task-randomized motion augmentation expands the executable skill space and improves downstream planning;
\textbf{(3)} whether staged training facilitates effective task learning and planner-level latent regularization improves motion quality while largely preserving task performance; and
\textbf{(4)} whether the learned policy transfers to a physical humanoid robot and supports sustained multi-skill rallies with human players.
We first describe the experimental setup, followed by baseline comparisons, augmentation analysis, training ablations, and real-world deployment.

\subsection{Experimental Setup}
\label{sec:exp_setup}

\paragraph{Simulation Environment.}
All simulation training and evaluation are conducted in mjlab~\citep{Zakka2026MJLab}.
We instantiate the system on the Unitree G1 humanoid robot with 29 degrees of freedom, although the proposed framework is not tied to this specific morphology.
The racket is modeled with explicit net, shaft, and grip collision geometries, with contacts resolved by the MuJoCo Warp physics backend and randomized surface friction.

\paragraph{Shuttle Initialization.}
During evaluation, shuttles are continuously served from the opponent side with randomized initial states.
The shuttlecock is modeled as a $5\,\mathrm{g}$ sphere for contact simulation, with quadratic aerodynamic drag governing its flight:
\begin{equation}
\dot{\mathbf{v}}=
\mathbf{g}
-k\|\mathbf{v}\|\mathbf{v},
\end{equation}
where $\mathbf{v}$ is the shuttle velocity, $\mathbf{g}$ denotes gravitational acceleration, and $k$ is a drag coefficient randomly sampled for each shuttle.
This produces incoming trajectories with varying speeds, heights, and lateral offsets, including high trajectories that create opportunities for jump returns.
Serves are launched every $1.5$--$2.0$ seconds.
The left panel of Fig.~\ref{fig:setup_latent} visualizes representative incoming trajectories.

\paragraph{Evaluation Metrics.}
We evaluate task performance using \textbf{Success Rate (SR)}, defined as the percentage of served shuttles successfully returned to the valid opponent court region.
Robot falls are counted as failures and trigger a reset.
Detailed success criteria are provided in Appendix~\ref{app:sim_eval_protocol}.
We further report \textbf{SR-F}, \textbf{SR-B}, and \textbf{SR-J}, defined as the percentages of all served shuttles successfully returned using forehand, backhand, and jump-return behaviors, respectively.
Each successful return belongs to exactly one category, such that
$$
\mathrm{SR}
=
\mathrm{SR\text{-}F}
+
\mathrm{SR\text{-}B}
+
\mathrm{SR\text{-}J}.
$$
\textbf{Average Consecutive Hits (AC)} measures rally stability as the average number of consecutive successful returns in completed streaks, capped at 10 in simulation.

To evaluate motion quality, we report \textbf{Joint Acceleration}, \textbf{Joint Jerk}, and \textbf{Joint Torque}. Lower acceleration and jerk indicate smoother motion, while lower torque indicates reduced average actuator effort.
We additionally report \textbf{JFID}~\citep{zhang2025natural}, which measures the distributional difference between policy-generated motions and the Stage~1 positive rollouts.
Derived from human motion references and adapted through physics-based RL, these rollouts provide a physics-grounded, executable motion prior.
Lower JFID indicates closer agreement with this motion prior, serving as a relative indicator of motion naturalness.
Since the reference distribution is generated by our own pipeline, we interpret JFID as a complementary metric rather than a method-agnostic measure.

\subsection{Comparison with Baselines}

We compare our full system with two representative baselines.
\textbf{Direct PPO} learns the badminton return task directly in the joint-action space without motion priors or hierarchical skill abstraction.
\textbf{AMP}~\citep{peng2021amp} incorporates an adversarial motion prior constructed from the same retargeted human badminton motions, while retaining joint-level task optimization.

Table~\ref{tab:baseline} summarizes the results.
Direct PPO achieves a reasonable level of return success but produces poor motion quality, exhibiting the highest joint acceleration, jerk, and torque, while failing to cover the full stroke repertoire.
AMP substantially improves motion smoothness and achieves the lowest joint jerk, but its behavior remains biased toward a limited set of skills, with no successful backhand returns; its joint acceleration and torque also remain higher than ours.
In contrast, our method achieves the highest overall success rate and rally stability while maintaining substantial success across forehand, backhand, and jump returns.
Despite this broader skill coverage, it obtains the lowest joint acceleration and torque, indicating less abrupt whole-body motion and reduced actuator effort.
It also achieves the lowest JFID, indicating closer agreement with the physics-grounded motion prior represented by the Stage~1 positive rollouts.
Together, these results demonstrate the effectiveness of planning in a target-conditioned latent skill space for achieving diverse and successful returns while maintaining motion quality.

\begin{table*}[t]
\centering
\caption{
Comparison with baseline methods.
SR-F, SR-B, and SR-J denote the percentages of all served shuttles successfully returned using forehand, backhand, and jump-return behaviors, respectively.
Lower Joint Acceleration, Joint Jerk, Joint Torque, and JFID indicate better motion-quality or distributional metrics.
}
\label{tab:baseline}
\resizebox{\textwidth}{!}{
\begin{tabular}{lccccccccc}
\toprule
\textbf{Method}
& \textbf{SR} (\%) $\uparrow$
& \textbf{SR-F} (\%)
& \textbf{SR-B} (\%)
& \textbf{SR-J} (\%)
& \textbf{AC} $\uparrow$
& \textbf{Joint Acc.} $\downarrow$
& \textbf{Joint Jerk} $\downarrow$
& \textbf{Torque} $\downarrow$
& \textbf{JFID} $\downarrow$ \\
\midrule
Direct PPO
& 79.0
& 0.0
& 25.1
& 53.9
& 4.14
& 59.63
& 26743.64
& 8.09
& 1226.88 \\

AMP
& 84.6
& 38.4
& 0.0
& 46.2
& 5.06
& 35.48
& \textbf{13420.18}
& 6.55
& 295.96 \\

Ours
& \textbf{88.3}
& 34.6
& 32.9
& 20.8
& \textbf{5.98}
& \textbf{34.28}
& 19318.24
& \textbf{6.13}
& \textbf{42.61} \\
\bottomrule
\end{tabular}
}
\end{table*}

\subsection{Effectiveness of Task-Randomized Motion Augmentation}
\label{sec:exp_aug}

\begin{figure}[t]
    \centering
    \includegraphics[
        width=0.9\linewidth,
        trim=0 230 20 220,
        clip
    ]{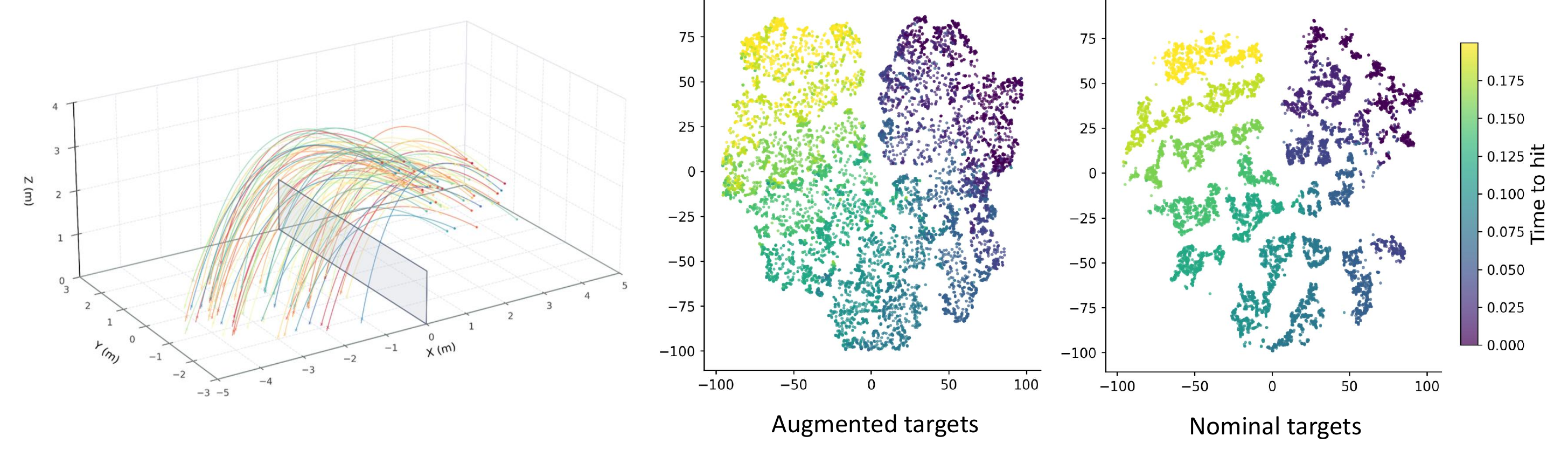}
    \caption{
    \textbf{Left:}
    Randomized shuttle initialization generates diverse incoming trajectories, including high trajectories that create opportunities for jump returns.
    \textbf{Right:}
    t-SNE visualization of target-conditioned latent codes collected from Stage~1 rollouts under a fixed reference motion and hitting event.
    Color indicates the remaining time to hit.
    Randomized-target codes spread around the corresponding nominal-target codes, forming local neighborhoods of executable stroke variations.
    }
    \label{fig:setup_latent}
\end{figure}

We evaluate whether task-randomized motion augmentation expands the executable skill space and improves downstream planning.
For visualization, we collect Stage~1 rollouts under nominal and randomized hitting targets for a fixed reference motion segment and annotated hitting event.
As shown in the right panel of Fig.~\ref{fig:setup_latent}, nominal-target latent codes form compact structures along the time-to-hit progression, whereas randomized-target codes spread around the corresponding nominal codes.
Because the skill encoder and low-level controller are jointly trained to realize these randomized racket targets, this spread represents executable target-conditioned stroke variations rather than arbitrary responses in latent space.
Task-randomized motion augmentation therefore expands each annotated hitting event into a local neighborhood of executable skills. 

We further train high-level planners on low-level controllers learned with and without task-randomized augmentation while keeping the remaining training setup unchanged.
As shown in Table~\ref{tab:augmentation}, removing augmentation substantially reduces return performance and rally stability, with no successful backhand returns.
These results suggest that the limited retargeted references alone do not provide sufficient skill variation to accommodate diverse incoming shuttle trajectories.
By adapting the reference motions to randomized racket targets, task-randomized augmentation provides a broader set of executable stroke variations for high-level planning, substantially improving task performance and multi-skill coverage. 
Additional analysis of how the Stage~2 planner organizes and composes continuous latent skills is provided in Appendix~\ref{app:latent_analysis}.

\begin{table}[t]
\centering
\caption{
Ablation on task-randomized motion augmentation.
SR-F, SR-B, and SR-J decompose the overall success rate into forehand, backhand, and jump-return behaviors, respectively.
}
\label{tab:augmentation}
\resizebox{0.7\linewidth}{!}{
\begin{tabular}{lccccc}
\toprule
\textbf{Method}
& \textbf{SR} (\%) $\uparrow$
& \textbf{SR-F} (\%)
& \textbf{SR-B} (\%)
& \textbf{SR-J} (\%)
& \textbf{AC} $\uparrow$ \\
\midrule
w/o Task-Rand. Aug.
& 57.0
& 40.2
& 0.0
& 16.8
& 2.38 \\
Stage 2 Only
& \textbf{89.2}
& 24.5
& 25.6
& 39.1
& \textbf{5.92} \\
\bottomrule
\end{tabular}
}
\end{table}

\subsection{Ablation Studies on Training Pipeline and Regularization}
\label{sec:exp_ablation}

We ablate two key design choices: the staged training pipeline and planner-level latent skill regularization.

\paragraph{Staged Training.}
We first evaluate whether task learning should be separated from planner-level latent skill regularization.
The \textbf{w/o Staged Training} variant applies the task and regularization rewards jointly from the beginning of planner training, whereas our staged pipeline introduces regularization only after task competence is established in Stage~2. 
As shown in Table~\ref{tab:staged_training}, applying planner-level regularization from the beginning substantially degrades return performance, rally stability, and stroke diversity.
This suggests that imposing regularization before reliable return behaviors are learned restricts exploration of the latent skill space.
In contrast, the staged pipeline first establishes task competence through latent-space exploration and then regularizes planner-level skill usage, preserving strong task performance and broad stroke coverage.

\begin{table}[t]
\centering
\caption{
Ablation on the staged training pipeline.
}
\label{tab:staged_training}
\resizebox{0.7\linewidth}{!}{
\begin{tabular}{lccccc}
\toprule
\textbf{Method}
& \textbf{SR} (\%) $\uparrow$
& \textbf{SR-F} (\%)
& \textbf{SR-B} (\%)
& \textbf{SR-J} (\%)
& \textbf{AC} $\uparrow$ \\
\midrule
w/o Staged Training
& 55.9
& 55.1
& 0.0
& 0.8
& 2.42 \\

Ours
& \textbf{88.3}
& 34.6
& 32.9
& 20.8
& \textbf{5.98} \\
\bottomrule
\end{tabular}
}
\end{table}

\paragraph{Planner-Level Latent Skill Regularization.}
We next evaluate the effect of Stage~3 regularization by comparing the Stage~2 planner with the full method.
Stage~2 already achieves strong task performance and substantial success across all three return types, confirming that the target-conditioned latent skill space supports effective high-level planning.
However, task-driven optimization alone does not explicitly constrain how the planner uses the learned latent skills.
As shown in Table~\ref{tab:latent_regularization}, Stage~3 substantially reduces JFID and joint jerk while also lowering joint acceleration and torque, maintaining comparable overall return performance and rally stability.
These results indicate that planner-level latent skill regularization improves motion quality and brings planner rollouts closer to the physics-grounded motion prior while preserving task competence and coverage of all three return types.

\begin{table*}[t]
\centering
\caption{
Effect of planner-level latent skill regularization.
}
\label{tab:latent_regularization}
\resizebox{\textwidth}{!}{
\begin{tabular}{lccccccccc}
\toprule
\textbf{Method}
& \textbf{SR} (\%) $\uparrow$
& \textbf{SR-F} (\%)
& \textbf{SR-B} (\%)
& \textbf{SR-J} (\%)
& \textbf{AC} $\uparrow$
& \textbf{Joint Acc.} $\downarrow$
& \textbf{Joint Jerk} $\downarrow$
& \textbf{Torque} $\downarrow$
& \textbf{JFID} $\downarrow$ \\
\midrule
Stage 2 Only
& \textbf{89.2}
& 24.5
& 25.6
& 39.1
& 5.92
& 34.62
& 23125.80
& 7.05
& 205.65 \\

Ours
& 88.3
& 34.6
& 32.9
& 20.8
& \textbf{5.98}
&  \textbf{34.28}
& \textbf{19318.24}
&  \textbf{6.13}
& \textbf{42.61} \\
\bottomrule
\end{tabular}
}
\end{table*}

\subsection{Real-World Deployment}
\label{sec:exp_real}

We finally evaluate sim-to-real transfer by directly deploying the simulation-trained policy on the physical humanoid robot.
During deployment, a motion-capture system provides real-time shuttle and robot base positions, while onboard sensors provide proprioceptive observations.
The physical badminton net is omitted for safety and motion-capture visibility.

We conduct 20 consecutive human--robot rally trials.
In each trial, a human player continuously returns the shuttle, and the robot responds using the learned policy.
All trials are recorded using a camera, and each robot return is manually labeled as a successful forehand return, backhand return, jump return, or failure.
Human-side misses are ignored in the statistics, whereas robot-side misses and falls are counted as failures.
Consecutive-hit statistics are reported without a cap.

\begin{figure}[ht]
\centering
\includegraphics[
width=0.7\linewidth,
trim=250 200 250 220,
clip
]{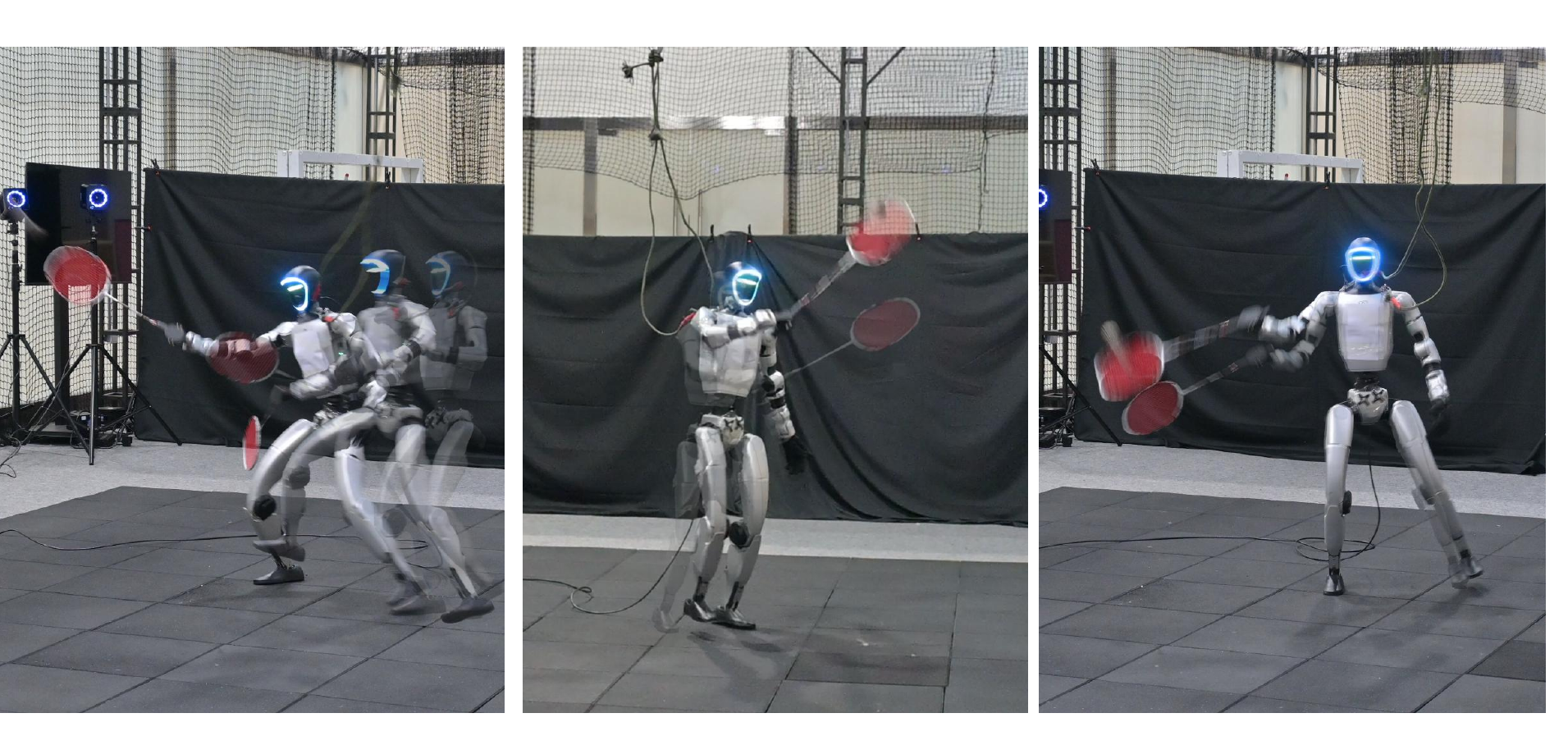}
\caption{Real-world executions featuring diverse strokes and coordinated whole-body movement.}
\label{fig:real_world}
\end{figure}

As shown in Table~\ref{tab:real_world}, the robot achieves an overall success rate of 89.4\%, with successful forehand, backhand, and jump returns, showing that the learned policy transfers to hardware without collapsing to a single return mode.
The system also sustains multi-shot human--robot rallies, averaging 8.42 consecutive successful returns and reaching a maximum streak of 23.
Representative real-world executions are shown in Fig.~\ref{fig:real_world}, with additional videos available on our \href{https://sunlight02.github.io/humanoid-badminton/}{project website}.

\begin{table}[t]
\centering
\caption{
Real-world deployment results.
Max CH denotes the maximum number of consecutive successful returns. Consecutive-hit statistics are uncapped.
}
\label{tab:real_world}
\resizebox{0.7\linewidth}{!}{
\begin{tabular}{lcccccc}
\toprule
\textbf{Setting}
& \textbf{SR} (\%) $\uparrow$
& \textbf{SR-F} (\%)
& \textbf{SR-B} (\%)
& \textbf{SR-J} (\%)
& \textbf{AC} $\uparrow$
& \textbf{Max CH} $\uparrow$ \\
\midrule
Real World
& 89.4
& 60.2
& 14.2
& 15.0
& 8.42
& 23 \\
\bottomrule
\end{tabular}
}
\end{table}

\section{Limitations}

Our system has two main limitations.
First, it focuses on returning the shuttle without explicitly controlling its landing location, limiting strategic shot placement.
Incorporating controllable return targets could enable more tactical play.
Second, real-world deployment relies on motion capture for shuttle and robot base states.
Replacing these inputs with onboard perception and state estimation is an important direction for future work.

\section{Conclusion}

We presented a three-stage hierarchical learning framework for humanoid badminton that enables dynamic racket skill learning from limited human motion data.
Through task-randomized motion augmentation, latent-space planning, and planner-level latent skill regularization, our method achieves improved task performance, stroke diversity, and motion quality over baseline methods.
Real-world deployment further demonstrates sustained multi-skill badminton rallies, including forehand, backhand, and highly dynamic jump returns, offering a step toward more agile and interactive humanoid control.

\clearpage

\acknowledgments{
This work was supported in part by the InnoHK initiative of the Innovation and Technology Commission of the Hong Kong Special Administrative Region Government via the Hong Kong Centre for Logistics Robotics.
This work was also supported in part by project MMT-p8-26 of the Shun Hing Institute of Advanced Engineering, The Chinese University of Hong Kong.
The authors thank Junhao Huang, Yueyang Wang, and Jiamu Qin for their assistance in collecting the badminton motion data.
}

\bibliography{corl_2026_template_submission/main}  

\clearpage

\appendix

\section{Training Details}
\label{app:training_details}

\subsection{Motion Processing and Task-Randomized Motion Augmentation}
\label{app:motion_processing}

We use approximately 30 minutes of human badminton videos as the motion source. 
The videos are divided into short segments of up to $10~\mathrm{s}$, each containing one or multiple hitting events. We reconstruct human motion from monocular video using GVHMR~\citep{Shen2024GVHMR}, represent it with SMPL~\citep{Loper2015SMPL}, and retarget it to the 29-DoF Unitree G1 humanoid using GMR~\citep{Araujo2025GMR}, obtaining reference joint trajectories $q_{\mathrm{ref}}$.

Since the videos do not provide explicit shuttle states, each hitting event is described by racket kinematics at an annotated hit frame $t_k^*$. We define its hit window as
\begin{equation}
    \mathcal{W}_k
    =
    [t_k^*-\delta,t_k^*+\delta],
\end{equation}
where $\delta=2$, corresponding to five frames. Multiple hitting events within a segment are processed independently, and $\mathcal{W}$ denotes the union of their hit windows.

For each hitting event, we sample a target racket state
\begin{equation}
    \tau_{\mathrm{target}}
    =
    (p_{\mathrm{target}},
     v_{\mathrm{target}},
     d_{\mathrm{target}}),
\end{equation}
comprising the target contact position, racket velocity, and racket-face direction. Targets are resampled online during training without modifying the reference trajectories. Both the sampled target and hit timing are provided to the skill encoder before contact; the hit window determines only when target-matching rewards are applied, while reference tracking constrains the surrounding motion.

The target contact position is sampled around the reference racket position in the world frame:
\begin{equation}
    p_{\mathrm{target}}
    =
    p_{\mathrm{racket}}(t_k^*)+\Delta p,
    \qquad
    \Delta p
    \sim
    \mathcal{U}
    \left(
    [-0.10,-0.10,-0.05],
    [0.10,0.10,0.05]
    \right).
\end{equation}
The target racket speed is sampled from $[3.0,6.0]~\mathrm{m/s}$, with a yaw offset of $\pm30^\circ$ around the court-forward direction. 
The nominal pitch is linearly interpolated from $45^\circ$ at a contact height of $1.0~\mathrm{m}$ to $0^\circ$ at $2.2~\mathrm{m}$, with values outside this range clipped to the boundary values, and an additional perturbation of $\pm30^\circ$.
The racket-face direction is sampled within $30^\circ$ of the target velocity direction. For the latent-space visualization, the nominal target uses the reference racket position, a speed of $4.5~\mathrm{m/s}$, and zero additional directional perturbations.

\subsection{Low-Level Controller Training}
\label{app:low_level_training}

Stage~1 jointly trains a skill encoder and a low-level controller using PPO with a control frequency of $50\,\mathrm{Hz}$. The skill encoder is a three-layer MLP that takes reference-motion information, hit timing, the sampled target racket state, and robot context as input, and outputs an 8-dimensional latent skill code with a $\tanh$ output activation:
\begin{equation}
    z_t \in \mathcal{Z}=[-1,1]^8.
\end{equation}
The low-level controller maps the latent code and robot context $\tilde{s}_t=[s_t,a_{t-1}]$ to joint-position targets:
\begin{equation}
    a_t=\pi_\phi(z_t,\tilde{s}_t).
\end{equation}

The Stage~1 reward combines reference tracking outside the hit windows, target matching within them, and regularization:
\begin{equation}
    r_t =
    \mathbb{I}[t\notin\mathcal{W}]r_t^{\mathrm{track}}
    +
    \mathbb{I}[t\in\mathcal{W}]r_t^{\mathrm{target}}
    +
    w_{\mathrm{reg}}r_t^{\mathrm{reg}}.
\end{equation}
The tracking reward comprises whole-body imitation terms for the global root position and orientation, root linear velocity, joint positions, relative body positions and orientations, and body linear and angular velocities.
The target reward evaluates racket position, linear velocity, and face-direction errors relative to the sampled target.
The regularization terms penalize abrupt action changes, joint-limit violations, self-collisions, and overly rapid stepping.
After Stage~1 training, the low-level controller is frozen for the subsequent stages.

\subsection{High-Level Planner Training}
\label{app:planner_training}

With the low-level controller frozen, the Stage~2 planner is trained to output latent skill codes for the badminton return task.
At each policy step, the planner produces $z_t\in\mathcal{Z}$, which is executed by the frozen low-level controller.
The planner observes the robot context $\tilde{s}_t$, the five most recent shuttle positions, the robot base position, and the base heading direction.
Shuttle positions are represented in both the global court frame and a robot-centric yaw-aligned frame, providing court-level trajectory information and relative shuttle motion.

The physics simulation runs at $1000\,\mathrm{Hz}$, while the high-level planner and low-level controller both operate at $50\,\mathrm{Hz}$ with an action decimation factor of $20$.
Each training episode lasts $10\,\mathrm{s}$ and contains repeated randomized incoming shuttles from the opponent side.

The planner reward is organized into five functional groups, summarized in Table~\ref{tab:planner_reward_summary}.

\begin{table}[h]
\centering
\caption{Summary of reward groups used for high-level planner training.}
\label{tab:planner_reward_summary}
\small
\begin{tabular}{p{2.5cm} p{9.5cm}}
\toprule
\textbf{Type} & \textbf{Description} \\
\midrule

Pre-contact
& Encourages lateral alignment with the incoming shuttle, reduces the racket--shuttle distance, and aligns the pre-contact racket velocity with a feasible hitting direction. \\

Impact
& Encourages sufficient racket velocity and racket-face alignment with the desired return direction at first racket--shuttle contact. \\

Post-contact
& Encourages the returned shuttle to travel toward the opponent side with sufficient forward velocity. \\

Jump return
& Encourages elevated contact, upward whole-body motion, and aerial striking for suitable high shuttles, while penalizing unnecessary jumps. \\

Regularization
& Encourages upright body orientation and sufficient body height, while penalizing prolonged racket--shuttle contact, abrupt action changes, joint-limit violations, self-collisions, and overly rapid stepping. \\

\bottomrule
\end{tabular}
\end{table}

\subsection{Planner-Level Latent Skill Regularization}
\label{app:latent_adv}

For Stage~3, we collect positive samples by rolling out the trained Stage~1 skill encoder and low-level controller under randomized hitting tasks, obtaining approximately $10$ million latent--context pairs. Negative samples are generated online from the high-level planner during adversarial training.

The context-conditioned discriminator $D_\psi$ takes $(z_t,\tilde{s}_t)$ as input and predicts the probability that the pair comes from the Stage~1 positive rollout distribution. It is trained using binary cross-entropy with a positive-label target of $0.9$ and a gradient penalty on positive samples with coefficient $1.0$. Its output defines the planner-level latent-style reward:
\begin{equation}
    r_t^{\mathrm{style}}
    =
    -\log\left(
        1-D_\psi(z_t,\tilde{s}_t)
    \right).
\end{equation}

The Stage~3 planner is initialized from the Stage~2 checkpoint and optimized using the original task reward together with the latent-style reward:
\begin{equation}
    r_t^{\mathrm{total}}
    =
    r_t^{\mathrm{task}}
    +
    w_{\mathrm{style}}r_t^{\mathrm{style}},
\end{equation}
where $w_{\mathrm{style}}$ balances task performance and planner-level latent skill regularization. The low-level controller remains frozen throughout this stage.

\section{Evaluation Details}
\label{app:evaluation_details}

\subsection{Simulation Evaluation Protocol}
\label{app:sim_eval_protocol}

For each policy, simulation metrics are computed over approximately $1{,}000$ randomized served shuttles (possibly slightly exceeding this number due to parallel evaluation).
A serve is terminated when the shuttle reaches the landing height, exceeds its sampled time limit, or the robot falls.
A robot fall counts the current serve as a failure and triggers a reset.

A serve is counted as successful if the shuttle makes contact with the racket, passes over the net, and lands within the valid opponent court region.
The valid landing region has a longitudinal length of $6.7\,\mathrm{m}$ and a singles width of $5.18\,\mathrm{m}$.
Since the simulated Unitree G1 humanoid is shorter than an adult human player, we set the net height to $1.2\,\mathrm{m}$ in simulation.

The overall success rate is defined as
\begin{equation}
    \mathrm{SR}
    =
    \frac{N_{\mathrm{success}}}{N_{\mathrm{serve}}} \times 100\%,
\end{equation}
where $N_{\mathrm{serve}}$ is the total number of served shuttles and $N_{\mathrm{success}}$ is the number of successful returns.

Successful returns are classified as forehand, backhand, or jump returns based on the first racket--shuttle contact.
A return is classified as a jump return if both feet are airborne at contact; otherwise, it is classified as forehand or backhand according to the side of the racket plane on which the shuttle lies.
Failed serves are not assigned to a stroke category, since the intended stroke type may be ambiguous when no valid return is produced.

Let $N_F$, $N_B$, and $N_J$ denote the numbers of successful forehand, backhand, and jump returns, respectively.
The corresponding success rates are
\begin{equation}
    \mathrm{SR\text{-}X}
    =
    \frac{N_X}{N_{\mathrm{serve}}} \times 100\%,
    \qquad X\in\{F,B,J\}.
\end{equation}
Each successful return belongs to exactly one category, such that
\begin{equation}
    \mathrm{SR}
    =
    \mathrm{SR\text{-}F}
    +
    \mathrm{SR\text{-}B}
    +
    \mathrm{SR\text{-}J}.
\end{equation}
Average Consecutive Hits (AC) averages the lengths of nonempty sequences of consecutive successful returns, each ending upon a failure or reset. In simulation, the robot resets after 10 consecutive successes. Sequences still ongoing when evaluation ends are excluded.

\subsection{Motion Metrics}
\label{app:motion_metrics}
We report Joint Acceleration, Joint Jerk, and Joint Torque as complementary measures of motion quality and actuator effort.
Let $q_{t,i}$ and $\tau_{t,i}$ denote the position and applied torque of actuated joint $i$ at control timestep $t$, respectively, and let $d$ be the number of actuated joints.

Joint Acceleration is defined as the mean absolute joint acceleration over $T$ evaluated timesteps:
\begin{equation}
    \mathrm{Joint\ Acceleration}
    =
    \frac{1}{Td}
    \sum_{t=1}^{T}
    \sum_{i=1}^{d}
    \left|\ddot{q}_{t,i}\right|.
\end{equation}

Joint Jerk is estimated from the joint-position sequence
using the third-order finite difference:
\begin{equation}
    j_t
    =
    \frac{
        q_{t+3}
        -3q_{t+2}
        +3q_{t+1}
        -q_t
    }{\Delta t^3},
\end{equation}
where $\Delta t$ denotes the control timestep and $q_t$ is the joint-position vector.
Joint Jerk is the mean magnitude over valid timesteps:
\begin{equation}
    \mathrm{Joint\ Jerk}
    =
    \frac{1}{T-3}
    \sum_{t=1}^{T-3}
    \|j_t\|_2.
\end{equation}

Joint Torque is defined as
\begin{equation}
    \mathrm{Joint\ Torque}
    =
    \frac{1}{Td}
    \sum_{t=1}^{T}
    \sum_{i=1}^{d}
    |\tau_{t,i}|.
\end{equation}
Lower Joint Acceleration and Joint Jerk indicate less abrupt motion, while lower Joint Torque indicates reduced average actuator effort.

Following FID-based evaluation of joint-angle motion distributions~\citep{zhang2025natural}, we additionally report Joint FID (JFID) to measure the distributional difference between policy-generated motions and the Stage~1 positive rollout distribution.
Both motion sets are segmented into fixed-length clips, with each frame represented by the joint-position vector relative to the default configuration.
The clips are normalized using statistics from the Stage~1 positive rollouts.
We then apply a discrete cosine transform (DCT)~\citep{ahmed1974discrete} along the temporal dimension and retain the first $K$ low-frequency coefficients as clip-level features.

Gaussian distributions are fitted to the positive and policy feature sets, denoted by $(\mu_+,\Sigma_+)$ and $(\mu_\pi,\Sigma_\pi)$, respectively.
JFID is computed as
\begin{equation}
\begin{aligned}
\mathrm{JFID}
=
\|\mu_+-\mu_\pi\|_2^2
+
\mathrm{Tr}\left(
\Sigma_+
+
\Sigma_\pi
-
2\left(
\Sigma_+^{1/2}
\Sigma_\pi
\Sigma_+^{1/2}
\right)^{1/2}
\right).
\end{aligned}
\end{equation}
Lower JFID indicates closer agreement with the Stage~1 positive rollout distribution.

\section{Additional Experiments}
\label{app:additional_experiments}

\subsection{Target-Range Sensitivity}
\label{app:target_range_sensitivity}

We examine whether broader target randomization compromises reference-motion tracking or target-matching accuracy.
We consider three scales, $0.5\times$, $1\times$, and $2\times$ the default target-randomization range described in Appendix~\ref{app:motion_processing}.
We measure joint-tracking error outside the hit windows and racket position and velocity errors relative to the sampled targets.

Across the three settings, outside-window joint-tracking error remains nearly unchanged at $0.89$--$0.90$.
As the target range increases from $0.5\times$ to $2\times$, hit-position error increases from $7.1$ to $8.3\,\mathrm{cm}$, while target-velocity error increases from $0.89$ to $1.94\,\mathrm{m/s}$.
These results indicate that broader target randomization reduces target-matching precision without substantially degrading reference tracking outside the hit windows.

\subsection{Planner Latent-Space Analysis}
\label{app:latent_analysis}

\begin{figure*}[t]
\centering
\includegraphics[width=0.8\textwidth]{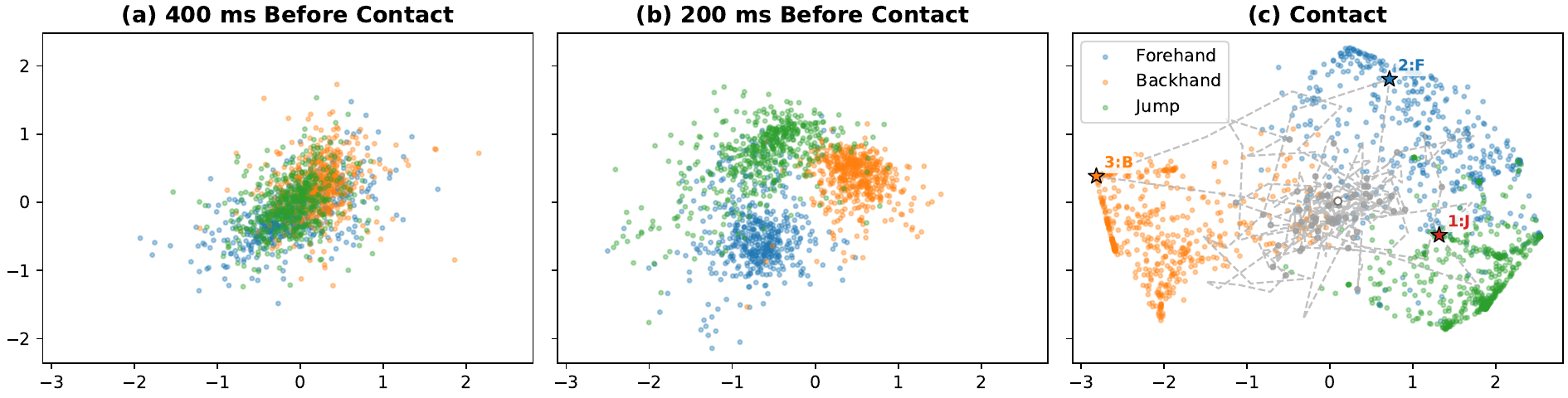}
\caption{
\textbf{Planner latent-space evolution.}
PCA visualization of successful forehand, backhand, and jump-return latent codes at
\textbf{(a)} $400$\,ms before contact,
\textbf{(b)} $200$\,ms before contact, and
\textbf{(c)} the contact frame, using a shared PCA basis.
The return types progressively separate as contact approaches.
The trajectory in (c) shows a consecutive three-return sequence covering all three return types; numbered markers indicate the return order and type.
}
\label{fig:latent_analysis}
\end{figure*}

We analyze how the Stage~2 planner organizes different return behaviors in the continuous latent skill space.
We project successful forehand, backhand, and jump-return latent codes onto a shared PCA basis at $400$\,ms and $200$\,ms before contact and at the contact frame.
As shown in Fig.~\ref{fig:latent_analysis}, the three return types largely overlap early in the return and progressively separate as contact approaches.
This suggests that the planner does not simply select fixed categorical skill codes, but instead continuously specializes its latent command according to the incoming shuttle and robot context.
At contact, we additionally overlay the latent trajectory from a consecutive three-return sequence covering all three return types.
The trajectory traverses different regions of the latent space across successive returns, illustrating how the planner composes continuous latent skills online during sustained rallies.

\end{document}